\documentclass[letterpaper, 10 pt, conference]{ieeeconf}  

\IEEEoverridecommandlockouts                           

\title{Learning Implicit Social Navigation Behavior using Deep Inverse Reinforcement Learning}

\author{Tribhi Kathuria, Ke Liu, Junwoo Jang, X. Jessie Yang, and Maani Ghaffari
\thanks{The authors are with the University of Michigan, Ann Arbor, MI 48109, USA.
    {\tt\footnotesize \{tribhi,kliubiyk,junwoo,xijyang,maanigj\}@umich.edu}}%
}

\usepackage[usenames,dvipsnames,svgnames,table,xcdraw]{xcolor}
\usepackage{amssymb} 
\usepackage{amsmath} 
\usepackage{graphicx} 
\usepackage{caption}
\usepackage{subcaption}
\usepackage{algorithm, algorithmicx, algpseudocode}

\usepackage[colorlinks=true,pdfpagemode=UseNone,citecolor=black,linkcolor=black,urlcolor=BrickRed]{hyperref}
\usepackage{cleveref} 
\usepackage{booktabs}
\usepackage{balance}

\begin{document}

\maketitle

\begin{abstract}


This paper reports on learning a reward map for social navigation in dynamic environments where the robot can reason about its path at any time, given agents' trajectories and scene geometry. Humans navigating in dense and dynamic indoor environments often work with several implied social rules. A rule-based approach fails to model all possible interactions between humans, robots, and scenes. We propose a novel Smooth Maximum Entropy Deep Inverse Reinforcement Learning (S-MEDIRL) algorithm that can extrapolate beyond expert demos to better encode scene navigability from few-shot demonstrations. The agent learns to predict the cost maps reasoning on trajectory data and scene geometry. The agent samples a trajectory that is then executed using a local crowd navigation controller.
We present results in a photo-realistic simulation environment, with a robot and a human navigating narrow crossing scenario. 
The robot implicitly learns to exhibit social behaviors such as yielding to oncoming traffic and avoiding deadlocks. We compare the proposed approach to the popular model-based crowd navigation algorithm ORCA and a rule-based agent that exhibits yielding. The code and dataset are publicly available at \mbox{{ \href{https://github.com/UMich-CURLY/habicrowd/tree/main}{https://github.com/UMich-CURLY/habicrowd}}.}
\end{abstract}
\section{Introduction}

The motion and behavior planning of robots are essential components of any real-world robotics stack. 
Classical planning approaches use simplistic map representations and model humans as static obstacles that are avoided using a reactive planner or controller. While these work well for controlled settings with uncomplicated scene geometry and simplistic interaction scenarios, this approach can fail in more complex task settings. 

Artificial Intelligence has recently seen huge advancements with sophisticated vision and language models applied to images, videos, and textual data sourced from internet-derived datasets\cite{CLIP, LatentDiffusionModel}. The Robotics community has since been looking to use these AI tools on robots, called embodied AI \cite{embodiedAIsurvey}. A prevalent learning paradigm for robots is Reinforcement learning, which aims to instill robotic agents with the capability to autonomously learn and adapt behaviors through interaction with their environment \cite{habitat1, RLsurvey}. 


\begin{figure}[t]
    \centering
   \begin{subfigure}{0.15\textwidth}
            \includegraphics[width=\textwidth]{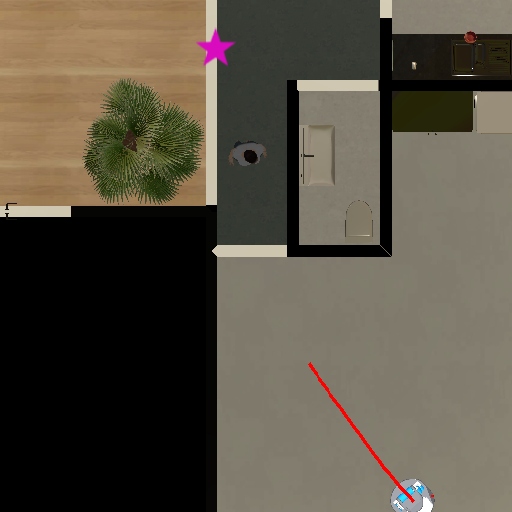}
            \caption{T = 0 sec}
            \label{fig:image1}
        \end{subfigure}
        \begin{subfigure}{0.15\textwidth}
            \includegraphics[width=\textwidth]{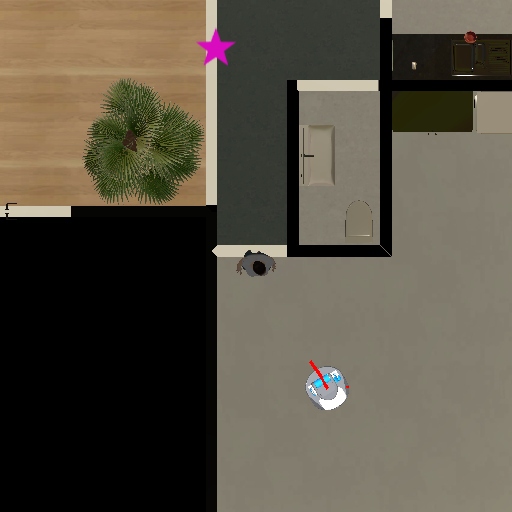}
            \caption{T = 2 sec}
            \label{fig:image2}
    \end{subfigure}
    \begin{subfigure}{0.15\textwidth}
            \includegraphics[width=\textwidth]{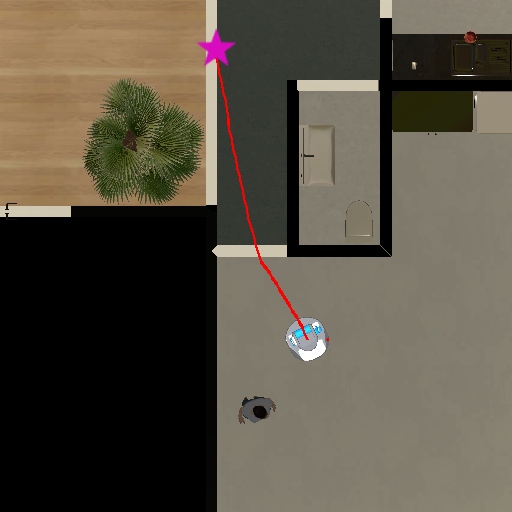}
            \caption{T = 3 sec}
            \label{fig:image3}
    \end{subfigure}
    \begin{subfigure}{0.15\textwidth}
            \includegraphics[width=\textwidth]{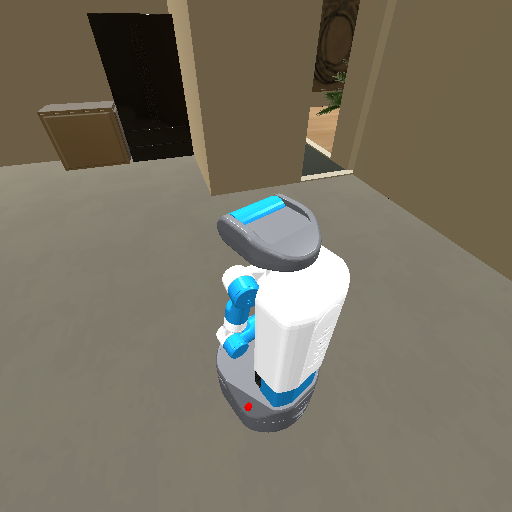}
            \caption{T = 0 sec}
            \label{fig:image4}
        \end{subfigure}
        \begin{subfigure}{0.15\textwidth}
            \includegraphics[width=\textwidth]{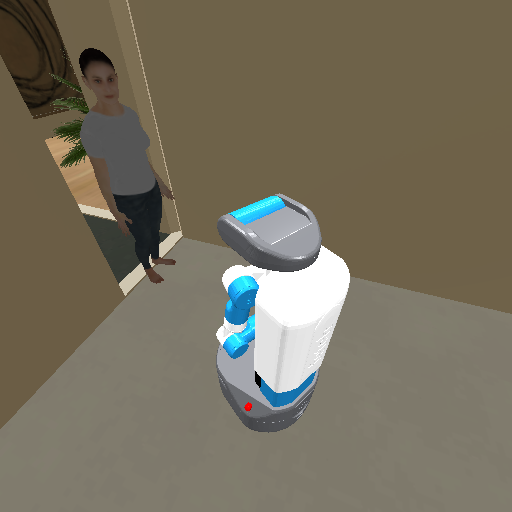}
            \caption{T = 2 sec}
            \label{fig:image5}
    \end{subfigure}
    \begin{subfigure}{0.15\textwidth}
            \includegraphics[width=\textwidth]{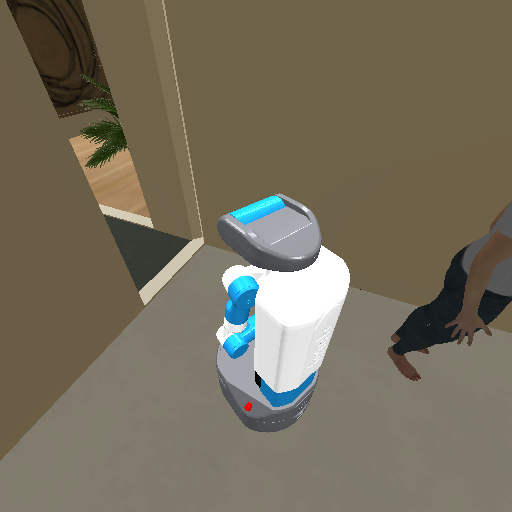}
            \caption{T = 3 sec}
            \label{fig:image6}
    \end{subfigure}

    \caption{Robots navigating in dense indoor environments with humans in the scene exhibit implicit social behaviors (legibility, yielding, and so on). 
    We learn from expert demonstration data to teach the robot these implicit navigation behaviors observed over a set of expert demonstrations. Figs.~\ref{fig:image1},~\ref{fig:image4} depict the beginning of an episode, where the expert decides a good position to be on their way to the goal while leaving room for the human. At the intermediary goal, the agent should likely wait for the human to pass and try to avoid it, as seen in Figs.~\ref{fig:image2},~\ref{fig:image4}. Finally, once the human has cleared the path, the agent decides to go to its goal, depicted by the Pink star in Figs.~\ref{fig:image3},~\ref{fig:image6}}
    \label{fig:demo}

\end{figure}

    



RL agents rely on handcrafted reward designs for task completion. In the context of social behaviors, these are usually ill-defined, sparse, or biased to design choice given the structure of the problem \cite{grid2pixilrl}.
While these behaviors are mainly intuitive to humans, it can be hard to encode them as well-defined rewards for RL agents. In social navigation, the most well-studied and benchmarked task is crowd navigation. \cite{CADRL} propose an RL agent trained on one such handcrafted reward design where the agent is given a large positive reward for successfully navigating to a goal and a small negative reward for getting too close to the dynamic obstacles or collisions. While this is a reasonable reward design and is able to complete the task in a controlled setting well, it is harder to generalize to more complicated task settings. Other improvements on this work, change the input or network design, which then improves performance given the chosen evaluation metrics, adding more bias to the training and evaluation. A perfect social RL agent requires us to theoretically model all interactive behaviors and design reward thresholds to see them in practice. 
This paper instead proposes an Inverse Reinforcement Learning (IRL) based planner that can learn rewards from a few shot expert demonstrations.  
IRL has been employed to recover rewards from demonstration successfully for many different tasks in robotics; these include highway lane changing scenarios \cite{liu2022inverse, prathiba2021intelligent}, crowd navigation \cite{fapl, solotdirl, kim2016socially}, and manipulation \cite{biyik2020active}. \cite{gan2022energy} use Maximum Entropy Deep Inverse Reinforcement Learning (MEDIRL) to learn traversability costmaps for legged robot locomotion. They also propose a trajectory ranking loss to encode a bias towards learning from demonstrations using less robot energy. 

We extend the MEDIRL approach to handle dynamic objects in the presence of complicated scene geometry. We extrapolate demonstration data to allow the agent to learn about image data with few-shot demonstrations \cite{brown2019extrapolating}. We can extrapolate static trajectories given a map of the environment and add that as extrapolated trajectory data for training the agent. To handle noise added by extrapolation, we propose a smoothing loss gradient that adds embedded navigation information given the robot's radius, similar to the inflation layer in an occupancy map~\cite{lu2014layered}. 
We choose an indoor narrow-crossing navigation scenario as a testbed for our agent as depicted in Fig.~\ref{fig:demo}. We train on raw sensor data obtained from the scene, and the agent learns to reason its path, understanding scene geometry and the underlying interaction dynamics implicitly. The agent learns to avoid deadlock and cause minimal disruption to the human trajectory. 
We compare our approach with a model-based crowd navigation algorithm, Optimal Reciprocal Collision Avoidance (ORCA), as well as a rule-based agent that yields to the human. We use a photo-realistic indoor environment in Habitat-Sim \cite{habitat3} to collect demonstration data using human experts and prepare a pipeline in Robot Operating System (ROS)\cite{quigley2009ros}. 


In particular, this work makes the following contributions:

\begin{enumerate}
\item We present a pipeline for training embodied agents to navigate dynamic, cluttered scenes using scene geometry and trajectory data extracted from top-down images. 
\item We propose an S-MEDIRL agent that has a smoothing loss for learning rewards and uses the extrapolation of demonstration using known scene geometry to learn static and dynamic obstacle avoidance from few-shot demonstrations. 
\item We present results in a narrow crossing environment and compare our performance with a state-of-the-art crowd navigation controller and a rule-based agent. 
\item We open-source our dataset and the code that can be used to generate other navigation scenarios using our setup at: \mbox{{ \href{https://github.com/UMich-CURLY/habicrowd/tree/main}{https://github.com/UMich-CURLY/habicrowd}}.}
\end{enumerate}

\section{Related Work}
\label{sec:related_work}
The following section discusses related work in social navigation using both learning and non-learning-based methodologies in Sections~\ref{sec: learning}, Section~\ref{sec: non_learning}. Finally, we introduce work in IRL to lay the foundation for our proposed methodology discussed in Section~\ref{sec: irl}

\subsection{Non-learning based Social Navigation}
\label{sec: non_learning}
Collision avoidance in human-robot interaction has been studied extensively. Some of the earlier methods were model-based. These included the social force model \cite{social_force} and its extensions, such as extended social force \cite{moussaid2010walking}, headed social force model \cite{farina2017walking}, and deep social force \cite{kreiss2021deep}, which use a proxemics-based model to avoid intruding on people's personal space in robot motion planning. Reciprocal velocity obstacle avoidance is another such approach where the robot builds sets of obstacles based on the observed velocity of other agents and then simultaneously solves for n-paths \cite{van2008reciprocal}. These have been adapted to account for different agent dynamics. Another popular reactive planning approach is dynamic window avoidance (DWA), an extension of the A* algorithm \cite{Fox1997TheDW}. However, a recurring problem with such reactive controllers is that they cannot resolve deadlock in a decentralized manner.

\cite{chandra2023decentralized} aims to resolve this deadlock in narrow crossing by separating the trajectory plan from the scheduling problem. They use a bi-level optimization algorithm that uses game-theoretic decentralized scheduling of agents given their different priorities (utilities). They, however, assume centralized control. In another work, \cite{thomas2018after} uses a rule-based agent to exhibit aggressive and assertive ways of resolving right of way through a door-crossing presented user study found that the assertive robot could resolve impasses. \cite{arul2023ds} model a terminal cost for two differential drive agents that helps avoid deadlock in their socially aware planner. While such model-based methods are explainable, they rely heavily on assumptions about interaction models and do not explicitly address scalability to multiple agents or heterogeneous environments.


\subsection{Deep Learning for Social Navigation}
\label{sec: learning}
Learning-based methods for social navigation have been primarily studied and benchmarked for the crowd navigation scenario \cite{mavrogiannis2023core}.  \cite{CADRL} proposes a deep reinforcement learning-based planner that can avoid collisions with other agents in the environment and minimizes its time to reach the goal simultaneously. The agent is trained in a $10 \times 10$ grid world with no obstacles, with other agents navigating using an ORCA model \cite{ORCA}. They define different multi-agent configurations that can cause traditional planners to freeze. Benchmarking this simplistic simulator setup, several studies have shown how to train a deep reinforcement learning agent for crowd navigation. 

Socially Aware Reinforcement Learning (SARL) adds a context of social norms to this cost formulation for crowd navigation \cite{sarl}. They explicitly model crowd dynamics on a coarse grid map and use that to anticipate human behavior for better robot planning. Similarly, \cite{DIPCAN} distill privileged human trajectory information in their training to have the agent implicitly learn to predict human motion and learn a navigation policy that can successfully avoid a collision. 
\cite{DZ} point out the inefficacy of the previously studied methods in deep reinforcement learning for crowd navigation to adapt to different human interaction modules. They define a safety critical reward characterized by a novel Danger One formalization that accounts for all possible human motion given different speeds. They improve the baseline methodologies in RL by adding this safety-critical reward defined on the danger zone. 

Other baselines work explicitly for crowd navigation; while most do not consider how to handle static obstacles in the scene, our previous work learns to navigate around humans in the presence of static obstacles \cite{solotdirl}. However, we do not consider scene dynamics; rather, we learn to reason about the static obstacle on the robot planning horizon. While this approach works well for a local planning problem, for successfully avoiding both static and dynamic obstacles in a scene, as we show in our results, these local planners need a reference from a planner that can reason about the scene geometry to handle more complicated environments. This paper presents one such case: the narrow crossing environment with a human walking from the other side, as depicted in Fig.~\ref{fig:demo}. For simplicity and computation time, we do not use a learning-based local planner for our approach, as is discussed in Section~\ref{sec:discussion}. 

\subsection{Inverse Reinforcement Learning}
\label{sec: irl}

A popular offshoot of Reinforcement Learning is Inverse Reinforcement learning. The IRL problem is a conjugate of the RL problem where we are given a set of expert demonstrations and need to find the reward function that best explains the demonstration data. For any given set of demonstrations, several rewards can explain human behavior. As a result, research has looked into optimally selecting these reward functions. \cite{ziebart2008maximum} proposed Maximum Entropy IRL (MaxEnt IRL) did feature matching under the constraint that the distribution over trajectories has maximum entropy. Other approaches for solving the IRL problem include Generative Adversarial Imitation Learning (GAIL) \cite{ho2016generativeadversarialimitationlearning}, Guided Cost Learning (GCL) \cite{finn2016guided}, Adversarial IRL (AIRL) \cite{fu2018learning}, and Bayesian IRL \cite{ramachandran2007bayesian}. 

The work of~\cite{wulfmeier2015maximum} generalizes the MaxEnt IRL approach to account for non-linear combinations of the features using neural networks, proposing Maximum Entropy Deep Inverse Reinforcement Learning (MEDIRL). 

The work of~\cite{lee2022spatiotemporal} uses IRL to train autonomous vehicles to merge in highway traffic. They propose a Goal-conditioned SpatioTemporal Zeroing (GSTZ)-MEDIRL framework that can zero out the reward for unvisited states. This is good for learning local control action and is suitable for highway driving, which does not have much dense static information to encode.

IRL has been used in a variety of applications, including manipulation \cite{finn2016guided}, autonomous vehicle control \cite{you2019advanced}, eye-tracking for attention detection \cite{baee2021medirl}, route planning \cite{zhao2023deep}, trajectory forecasts \cite{deo2020trajectory}. 

In the social navigation context, \cite{gao2019learning} learn how to approach groups of people using inverse reinforcement learning. The robot learns socially appropriate behavior toward human groups, ensuring natural and acceptable interactions. \cite{kim2016socially} use a wheelchair robot to learn to navigate a crowded corridor using a global planner to define a path through the corridor. \cite{solotdirl}, similarly use it for learning to navigate in crowded rooms, using a trajectory rank to learn more from demonstrations that have less jerk in human trajectories. Our work deals with learning social behavior based not only on dynamic agent trajectories but also on scene geometry. As a result, we learn to predict a reward map at any instant, given the images and trajectory data, as opposed to the handcrafted feature design in the previous works.







\section{Methodology}

\begin{figure*}[t]
    \centering
        \begin{subfigure}{0.55\textwidth}
            \includegraphics[width=\textwidth]{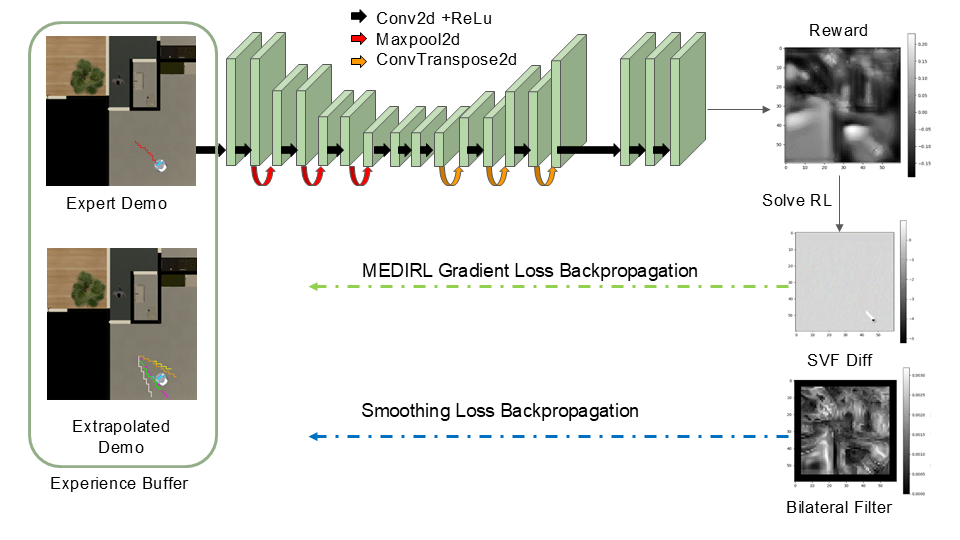}
    \caption{Offline Training}
    \label{fig:network-architecture}
        \end{subfigure}
        \begin{subfigure}{0.65\columnwidth}
            \includegraphics[width=\textwidth]{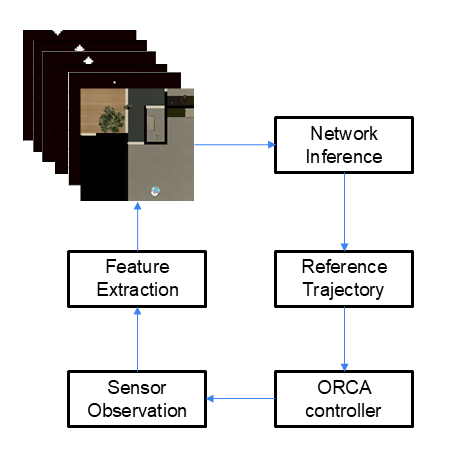}
            \caption{Online Deployment}
            \label{fig: inference}
        \end{subfigure}

    \caption{Fig.~\ref{fig:network-architecture} shows the training architecture. The expert demos are sampled every 0.1s to create a new demonstration data point. The past trajectories of the human and robot are fed in to give the network history of past positions. The U-Net architecture feeds into the final regression layer, which outputs the learned reward. The reward is then used to sample a trajectory ($E_\mu$), which is compared to the demonstrated trajectory at any time t, looking 10 steps in the future. This gradient is then backpropagated till the network converges. Fig.~\ref{fig: inference} shows the online deployment pipeline, where the reward inferred using the network is used to generate the reference trajectory for the local controller.}
    
\end{figure*}
\label{sec:methodology}

The proposed pipeline learns to navigate cluttered dynamic environments from expert demonstrations, reasoning about scene geometry and the trajectory of the agents in the scene. We present results on the narrow crossing scenario with one human and a robot trying to navigate from opposite sides and compare them to other model-based methods for crowd navigation. 
The following section explains our proposed S-MEDIRL algorithm and the data collection for the time-based implementation of the IRL agent to predict different rewards at any time $\mathrm{t}$.

\subsection {Markov Decision Process (MDP)}
\label{sec:irl}

The IRL agent learns the reward function sampling demonstration data from the expert demonstration buffer. We set the problem as an MDP with the following formulation:
\begin{itemize}
    \item \textbf{State space}: We use a low-resolution top-down image of the environment with grid size $60\times60$ pixels over $6\times6~\mathrm{m}^2$. The grid is centered on the door position, and the agent positions are known in the grid frame.
    \item \textbf{Action space}: The actions of the agent in the discretized state space are chosen to be \mbox{$\mathcal{A} = \{\text{up, down, left, right, stay}\}$}, which allow the robot to transit from the current grid point to an adjacent grid point. 
    \item \textbf{Transition function}: The transition between states is assumed to be deterministic. 
    
\end{itemize}

Given an MDP, a RL problem seeks the optimal policy \mbox{$\pi^\ast: \mathcal{S} \to \mathcal{A}$}, that maximizes the expected discounted reward similar to~\eqref{eq: RL_policy}. 
In the IRL problem, we have a set of Demonstrations ($\mathcal{D}$) on an MDP, and we want to recover the Reward function $R$. Where demonstrations are a set of observed trajectories $\tau$, which is a sequence of the observed state-action pairs $\tau = \{(s_1, a_1), (s_2, a_2), ..., (s_T, a_T)\}$. Given a set of features $\phi(s, a)$ that describe the state-action pairs, the reward function 
$R(s, a)$ can be represented as:
\begin{equation}
    R(s, a) = \theta^{\mathsf{T}}\phi(s, a)
    \label{eq: RL_policy}
\end{equation}

With this feature-matching algorithm, there are many possible solutions to the problem. Maximum entropy IRL resolves this reward ambiguity by working on the assumption that the demonstrations with higher rewards are exponentially more likely \cite{ziebart2008maximum}.

\subsection{MEDIRL}
MEDIRL~\cite{wulfmeier2015maximum} relaxes the assumption that the reward is a linear function of the features using neural networks to approximate the Reward function. 
\begin{equation}
    \mbox{$\hat{r}_{\theta}(s) = f(\phi(s);\theta)$} 
\end{equation}
The MEDIRL loss is then expressed as follows:
\begin{align}
\label{eq:medirl}
    \mathcal{L}(\theta) &= \log \mathbf{P}(\mathcal{D}, \theta | r_{\theta}) 
    = \log \mathbf{P} (\mathcal{D} | r_{\theta}) + \log \mathbf{P} (\theta) \nonumber \\
    &=: \mathcal{L}_\mathcal{D} + \mathcal{L}_{\theta}, 
\end{align}
Where  $\mathcal{L}_\mathcal{D}$ is the demonstration loss, $\mathcal{L}_{\theta}$ is the regularization loss. The loss gradient can then be expressed as follows: 
\begin{equation}
    \label{eq: gradient}
    \begin{aligned}
    \frac{\partial\mathcal{L}}{\partial\theta} &= \frac{\partial\mathcal{L}_{\mathcal{D}}}{\partial\theta} + \frac{\partial\mathcal{L}_{\theta}}{\partial\theta} 
    =\frac{\partial\mathcal{L}_{\mathcal{D}}}{\partial r} \cdot \frac{\partial r }{\partial\theta} + \frac{\partial\mathcal{L}_{\theta}}{\partial\theta} \\
    &= (\mu_{\mathcal{D}}-\mathbb{E}[\mu]) \cdot \frac{\partial r }{\partial\theta} + \frac{\partial\mathcal{L}_{\theta}}{\partial\theta} .
    \end{aligned}
\end{equation}

The first part of this sum can be computed as the difference between the expert and the sampled trajectory from the learned reward function, as can be seen in Fig.~\ref{fig:network-architecture}. The second is backpropagated as a regularization loss of the weights of the network.

\subsection{Smooth MEDIRL}

MEDIRL has been widely used to infer reward functions in static environments. This works well, as the agent's trajectory, given a scene, remains almost the same. As a result, the IRL agent can reason about scene geometry to infer characteristics such as traversability \cite{gan2022energy}. On the other hand, for the dynamic navigation task, the reward depends on the trajectory data of the agents. When navigating cluttered dynamic environments, both these problems get convolved together. 

\begin{figure}[h]

    \centering
   \begin{subfigure}{0.15\textwidth}
   \centering
            \includegraphics[width=\textwidth]{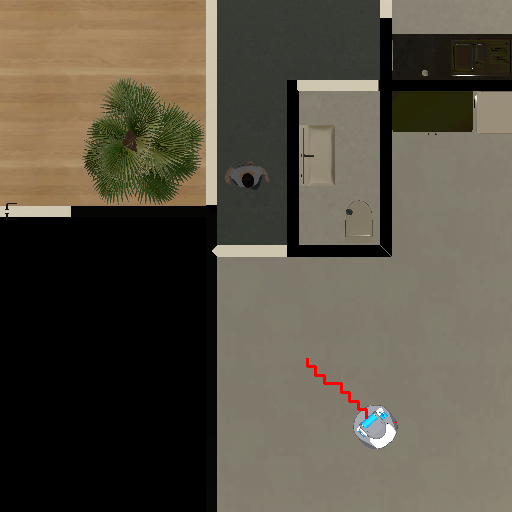}
            \caption{Expert demo}
            \label{fig: no_noise}
        \end{subfigure}
        \begin{subfigure}{0.15\textwidth}
        \centering
            \includegraphics[width=\textwidth]{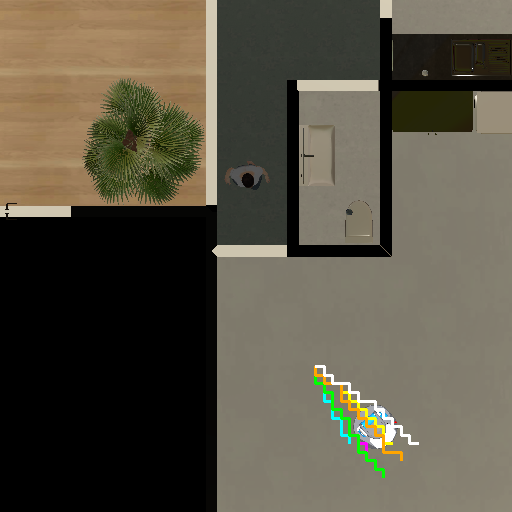}
            \caption{Extrapolated Demo}
            \label{fig: noise}
    \end{subfigure}
    \caption{The figure shows the augmented noise data fed into the network for one of our trajectories. Fig.~\ref{fig: no_noise} depicts the expert demo, while Fig.~\ref{fig: noise} depicts the added noisy trajectory data.}
    \label{fig: noise_overlay}
\end{figure}

To explicitly add navigability information to our network without increasing data collection overhead, we propose to extrapolate our demonstration data. The expert data consists of 15 demonstration data points collected over seven different start and goal positions of the robot and human in the scene. This is, however, insufficient to capture scene navigation as it leaves many states unseen; as a result, we extrapolate to our demonstration data depicted in Fig.~\ref{fig: noise_overlay}.
We extrapolate the current trajectory data by giving it six random start positions close to the current start of the robot. This helps the network generalize better to unseen test cases.

\begin{figure}[t]
    \begin{subfigure}{0.45\columnwidth} 
        \includegraphics[width=0.9\textwidth]{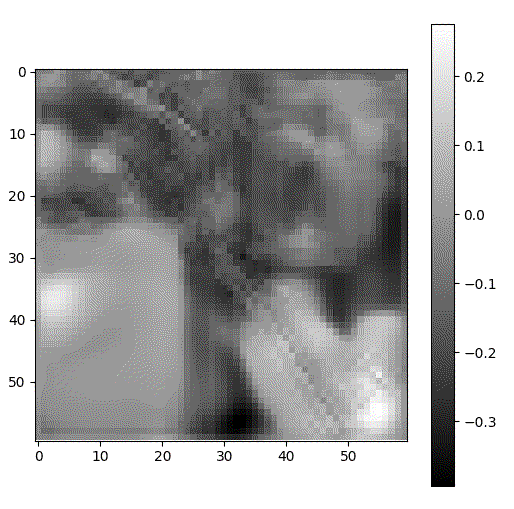} 
        \caption{Reward with smoothing}
        \label{fig:reward_0}
    \end{subfigure}
    \begin{subfigure}{0.45\columnwidth}
        \includegraphics[width=0.9\textwidth]{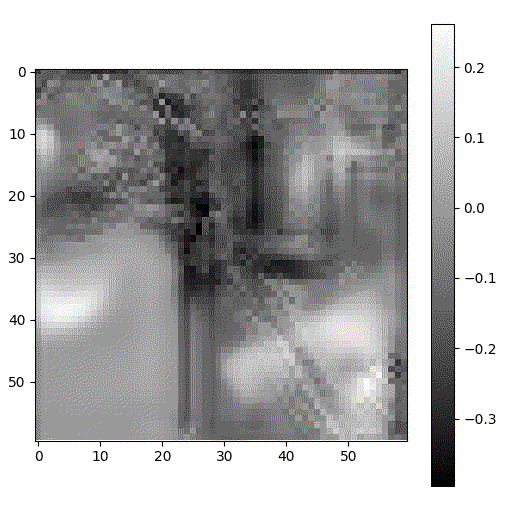} 
        \caption{Reward without smoothing}
        \label{fig:reward_1}
    \end{subfigure}

    \caption{Comparison of a reward at time $t$ trained with and without the smoothing loss. Fig.~\ref{fig:reward_0} shows that there are fewer pixels allocated starkly different values to its neighbors compared to Fig.~\ref{fig:reward_1}}
    \label{fig:reward_compare}
\end{figure}

To incorporate temporal information into the grid environment, explicit stopping points, or counter-stops, are added to the demonstration data. On a discrete grid with a fixed look-ahead, trajectories may appear to move even when the robot is stationary. To address this, we set a 3-second threshold to identify counter-stops, excluding states where the controller is merely turning in place. The reference trajectory spans 10 grid points, and data near counter-stops is augmented with stopping locations to help the robot learn to pause and wait as demonstrated. Fig.~\ref{fig: noise_overlay} shows the robot stopping at the first counter-stop.
Fig.~\ref{fig:reward_1} shows an instance of the robot navigating an unseen test case using the reward trained with noisy data. We note many added aberrations due to the added noise, where neighboring pixels have different reward values. The noise information being randomly sampled can bias the system to create some local minima and maxima in the reward.
We propose a smoothing loss that can inform the network of the scene traversability while eliminating the aberrations. The proposed loss, implemented as Bilateral Filtering loss, is applied to the trained reward values to ensure that the reward function is smooth and that the reward can generalize across similar states. Compared to the general robotics stack, this loss emulated the inflation layer of a cost map in our learning-based framework. 

The Bilateral filtering loss is defined in~\eqref{eq: new_loss} and is added to the MEDIRL loss given in~\eqref{eq:medirl}.
\begin{align}
\label{eq: new_loss}
\mathcal{L}_{\text{bilateral}}(r) &= \sum_{p} \sum_{q \in \mathcal{N}(p)} \left\| r(p) - r(q) \right\|^2 \exp \left( - \frac{\| p - q \|^2}{2\sigma_s^2} \right) \nonumber \\
&\quad \cdot \exp \left( - \frac{\| r(p) - r(q) \|^2}{2\sigma_c^2} \right).
\end{align}
Here, \(\mathcal{N}(p)\) represents the neighborhood of a pixel \(p\), and \(q\) is a neighboring point in \(\mathcal{N}(p)\). The terms \(\sigma_s\) and \(\sigma_c\) control the sensitivity to spatial distance (\(\|p - q\|\)) and residual difference (\(\|r(p) - r(q)\|\)), respectively, ensuring smoothness that respects both spatial and intensity variations.




Fig.~\ref{fig:reward_compare} shows the reward learned using the smoothing loss compared to the one learned without smoothing. There are fewer pixelated areas where the robot can be stuck in a local maxima. 

\subsection{Training and Online Deployment}
\begin{figure}[t]
    \begin{minipage}{0.55\textwidth}
        \begin{subfigure}{0.2\textwidth}
            \includegraphics[width=\textwidth]{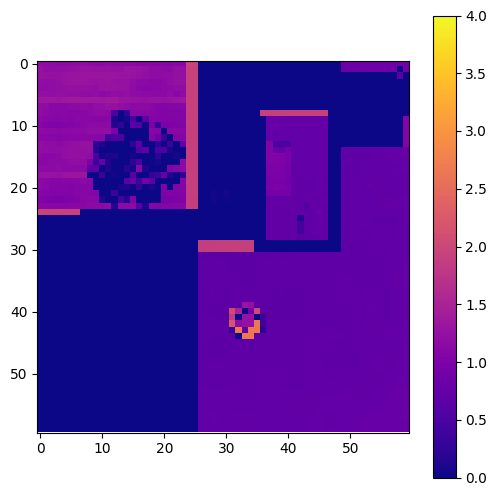}
            \caption{R Channel}
            \label{fig: rgb_r}
        \end{subfigure}
        \begin{subfigure}{0.2\textwidth}
            \includegraphics[width=\textwidth]{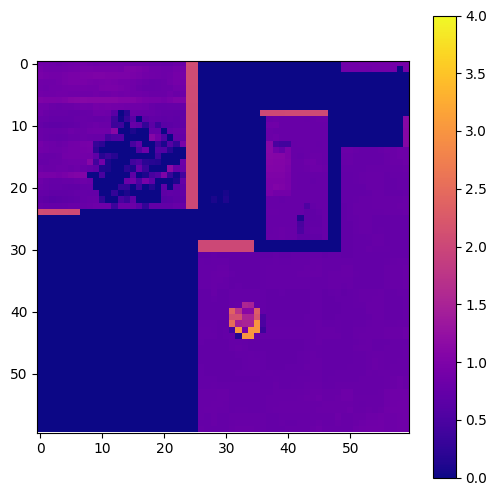}
            \caption{G Channel}
            \label{fig: rgb_g}
        \end{subfigure}
        \begin{subfigure}{0.2\textwidth}
            \includegraphics[width=\textwidth]{fig/101/heatmap_temp_r.png}
            \caption{B Channel}
            \label{fig: rgb_b}
        \end{subfigure}
        \begin{subfigure}{0.2\textwidth}
            \includegraphics[width=\textwidth]{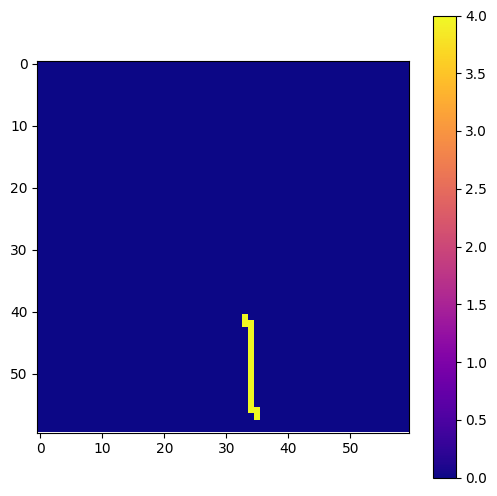}
            \caption{Robot past}
            \label{fig: robot_past}
        \end{subfigure}
    \end{minipage}
    \begin{minipage}{0.55\textwidth}
        \begin{subfigure}{0.2\textwidth}
            \includegraphics[width=\textwidth]{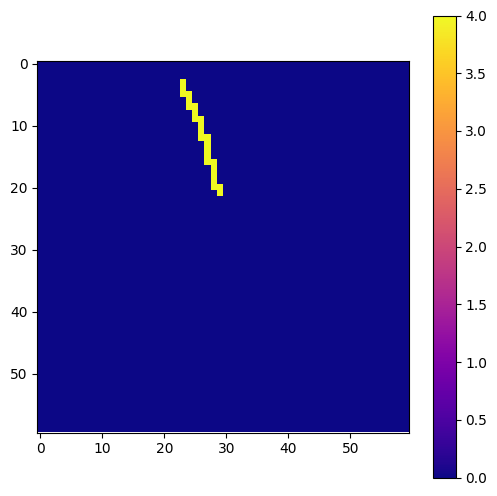}
            \caption{Human past}
            \label{fig: human_past}
        \end{subfigure}
        \begin{subfigure}{0.2\textwidth}
            \includegraphics[width=\textwidth]{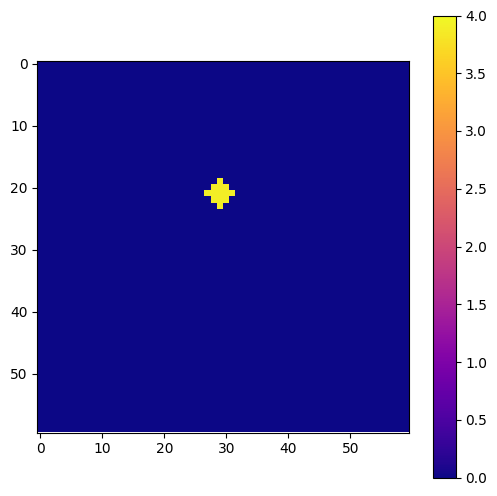}
            \caption{Heading}
            \label{fig: human_heading}
        \end{subfigure}
        \begin{subfigure}{0.2\textwidth}
            \includegraphics[width=\textwidth]{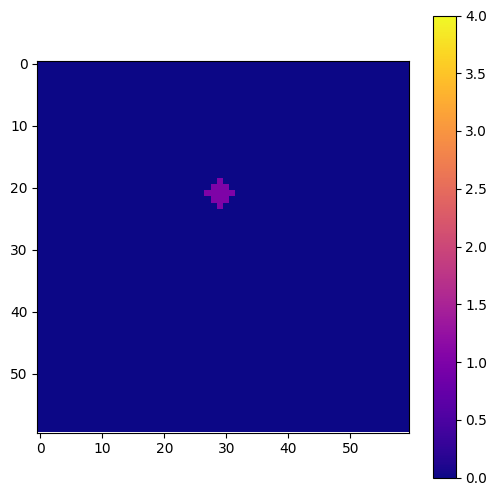}
            \caption{Velocity}
            \label{fig: human_vel}
        \end{subfigure}
        \begin{subfigure}{0.2\textwidth}
            \includegraphics[width=\textwidth]{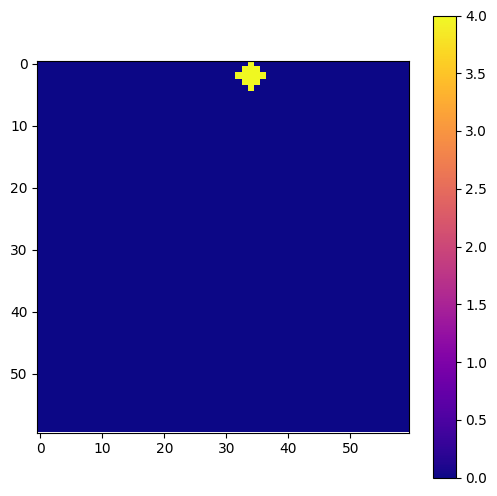}
            \caption{Robot Goal}
            \label{fig: robot_goal}
        \end{subfigure}
    \end{minipage}
    \caption{This figure provides a snapshot of the network's input features. The first three features, extracted from the RGB image of the demonstration's top-down view, are normalized and split into Red, Green, and Blue channels (Fig.~\ref{fig: rgb_r}-\ref{fig: rgb_b}). The network also receives the episode history, including the robot's past trajectory (Fig.~\ref{fig: robot_past}) and the human's past trajectory (Fig.~\ref{fig: human_past}). The human's current state, represented by velocity and heading (Fig.~\ref{fig: human_vel}-\ref{fig: human_heading}), and the robot's goal (Fig.~\ref{fig: robot_goal}) are also encoded as inputs.}
    \label{fig: network-inputs}
\end{figure}

The network input is a stacked feature set consisting of the images seen in Fig.~\ref{fig: network-inputs}. The features are collected every 0.1s apart, and the reference trajectory for each feature is extracted for 10 discrete steps in the future. The RGB image feature is divided into three normalized image features, each representing the RGB values respectively seen in Fig.~\ref{fig: rgb_r}-\ref{fig: rgb_b}. Other features are extracted from agents position; these include the human's past trajectory, seen in Fig.~\ref{fig: human_past}, and robots past trajectory in Fig.~\ref{fig: robot_past}. We also consider human dynamics such as heading and velocity depicted in Fig~\ref{fig: human_heading} and Fig.~\ref{fig: human_vel}, respectively. Finally, we condition the training to the robots goal as seen Fig.~\ref{fig: robot_goal}, which is assumed to be known, which is the case for most planning problems. 



As seen in Fig~\ref{fig: inference}, the network is then deployed to predict the reward given the top-down image and the human trajectory in the observed scene. The reward is used to then get a trajectory for the robot, resampled every 0.2 seconds. We use an ORCA controller to follow this reference. 
We present results on 13 unseen cases in the same scene as test results. We have a U-Net architecture depicted in Fig.~\ref{fig:network-architecture} similar to \cite{gan2022energy}. The robot learns to adapt to different start and goal positions of both the human and the robot and implicitly predicts how to yield to the human and when to start moving to the goal.

\section{Experiments}
\label{sec:experiments}
We present results collected in a Habitat-Sim simulation environment. We fix a scene setup and assign different start and goal positions to both the human and robot for each episode. For testing our learned agent, we have 13 unseen episodes and deploy the policy replanning the trajectory every 0.2s. The learned trajectory is then given to an ORCA controller that can avoid collisions between dynamic agents and return a velocity that the robot can execute. We use ROS to communicate between modules, convert images to feature space, infer sampled robot trajectory from predicted reward cost map, and finally, execute the reference trajectory using ORCA. 
\begin{figure*}[t]
    \centering
    \includegraphics[width=2.0\columnwidth]{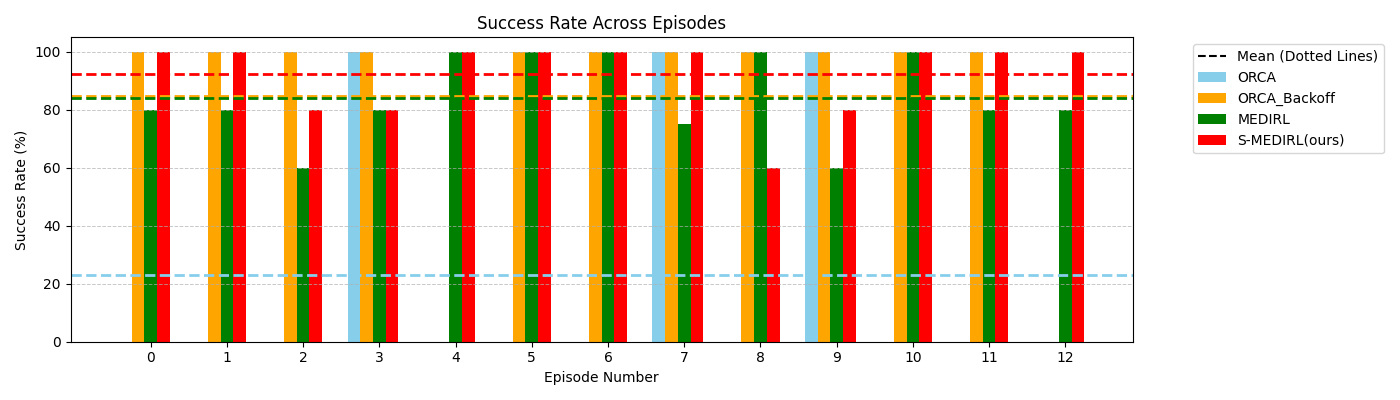}
    \caption{The figure shows the success rate of the MEDIRL agents compared to our model-based (ORCA) and rule-based (ORCA) baselines collected over 5 runs. For episodes that do not have deadlock and no negotiation involved in passing the narrow crossing (3, 7, and 9), all agents perform well and are able to complete the task. For episodes where the agents end up in deadlock, the nominal ORCA agent fails, but the backoff agent is able to succeed in most cases. The failure in the backoff case comes from the starting position not always being the best position to yield. The MEDIRL agent is prone to end up in local minima if it gets stuck in unseen states in the scene, and as result has a median success rate of about 82\%, similar to the ORCA backoff agent. While, our S-MEDIRL is able to successfully navigate the scene 92\% of the time. }
    \label{fig: success}
\end{figure*}
\subsection{Baselines}
We evaluate two baselines relevant to our task:

\subsubsection{ORCA Agent}
\label{orca_agent}
The first baseline is an Optimal Reciprocal Collision Avoidance (ORCA) agent, which uses a reference trajectory generated by a shortest path planner on a static map. This planner operates on navigation mesh edges, similar to a Voronoi planner, as detailed in \cite{kathuria2022providers}. While ORCA inherently avoids dynamic collisions, it relies on the higher-level planner to account for static obstacles. In our implementation, the ORCA agent replans every 0.2 seconds with a discrete step size of 1.33 m/s and a 4-second planning horizon. The maximum speed is 1.3 m/s, and dynamic agents are considered within a 0.5 m horizon. Static obstacles can occasionally constrain ORCA, causing deviations from its intended path.

\subsubsection{ORCA-Backoff Agent}
The second baseline introduces a rule-based backoff mechanism to address deadlocks. It builds upon the ORCA agent in Section~\ref{orca_agent} by adding a deadlock detection mechanismdeadlock is detected : if two agents move less than 0.1 m over 50 timesteps (~10 seconded. In such cases, the agent retreats to its start position, waits for the human agent to pass, and then proceeds.




\subsection{Results}
To test the performance of our IRL agent, we compare it against the baselines in a narrow crossing setup. The human agent is walking from the other side of the door and trying to pass the robot on the other side. The human agent does not back off and has priority in the path; the robot agent must either try to go to the goal or back off and yield to the human until the path is clear and then move towards its goal.

Fig.~\ref{fig: success} shows the rate of success of our agent (S-MEDIRL) compared to the baselines collected over 5 runs for each 13 episodes. The ORCA agent, getting its reference path from the shortest path planner, ends up in a deadlock for all the scenarios except episodes 3, 7, and 9. In case of no deadlock, this agent has a perfect success rate, as is expected. However, when there is a deadlock, the agents fail to complete the task. The ORCA backoff agent is able to complete most tasks except for 4 and 12. This agent is trained to choose the robot's start position to yield to the human, which in some cases, such as in episodes 4 and 12, are in the human's way, and it fails to let the human pass. The MEDIRL agent can successfully complete each scene but has some failure cases where it gets stuck in local minima or unseen states. The average success rate for this agent is similar to the Backoff agent, about 82\%. The S-MEDIRL agent can recover from most unseen states and bring up the average success rate to about 92\%. 

\begin{figure}
     \centering
     \includegraphics[trim={0 0 0 0}, clip, width=\columnwidth]{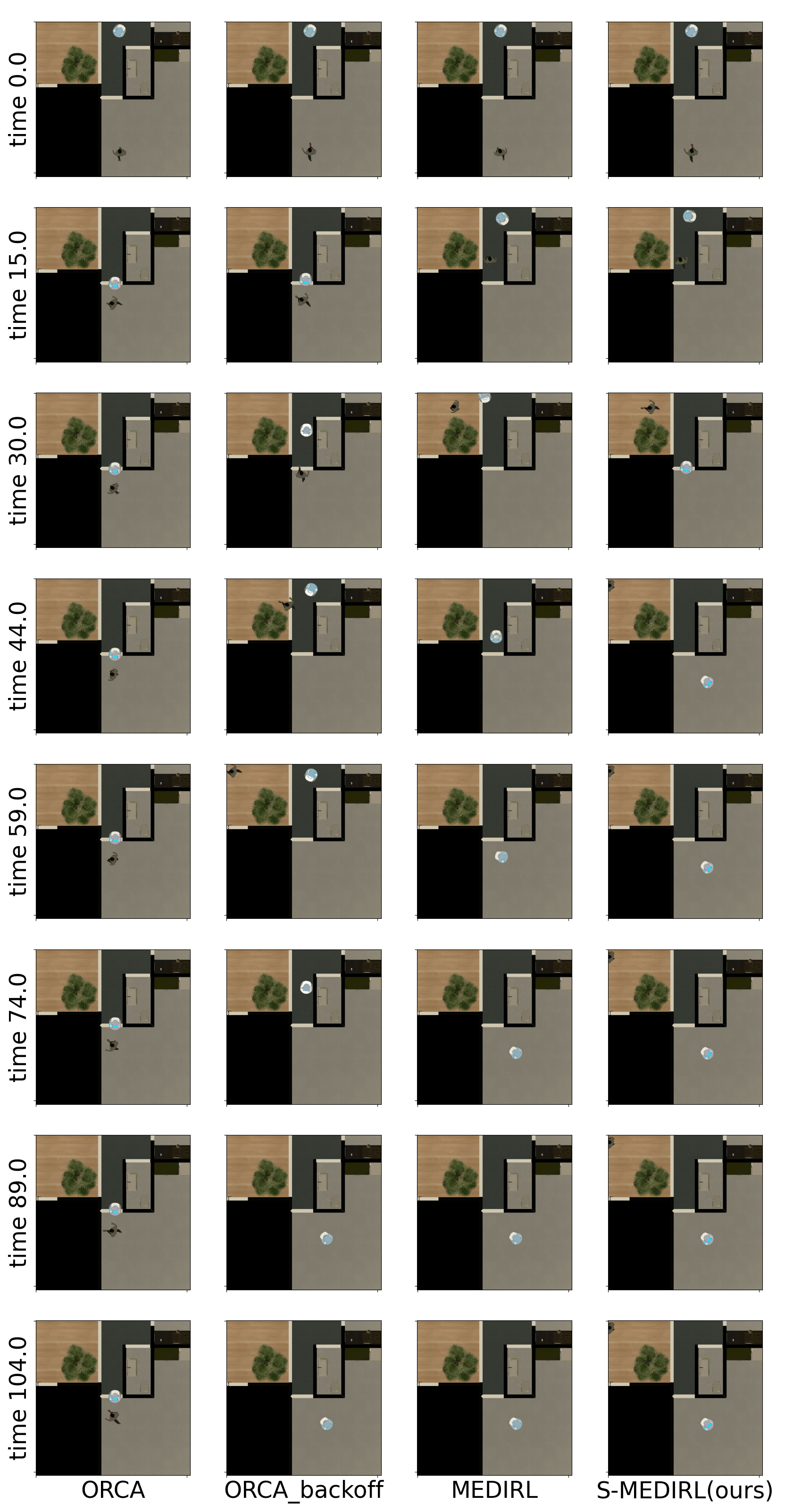}
     \caption{This figure shows snippets of the robot trajectory at discrete time instants. The agents position is noted, relative to the humans position and goal. The ORCA agent fails in the given scenario, and the agents end up in a deadlock. The ORCA-Backoff agent can successfully resolve the deadlock, but takes longer to complete the task, than the IRL agents. The MEDIRL is more prone to being stuck in local minima and ends up taking longer. }
     \label{fig:traj-time}
 \end{figure}

Qualitatively the agent's performance for a selected run of episode 8 can be seen in Fig.~\ref{fig:traj-time}. The S-MEDIRL agent is at the goal in under a minute, depicted in a discrete time of t = 44 seconds. The MEDIRL agents get stuck in a local minima for around t = 30 and end up taking longer to complete the task. 
The ORCA\_backoff agent finally completes the task at t = 89 seconds. As can be seen through the time-lapse, this agent first tries to go to the goal and end up in deadlock at t = 15 seconds. Then, it decides to back off and goes to its start position, seen in t = 30 to t = 59 seconds. Once the human has cleared at t = 59 seconds, the agent finally decides to move to the goal. The IRL agents instead avoid this deadlock by waiting for the human to pass first and leaving room for the human, as seen in t = 15s to t = 30s, and then start to move to its goal when the path is clear. The ORCA agent ends up in a deadlock and cannot finish the task.

\begin{figure}
    \begin{subfigure}{\columnwidth} 
    \centering
        \includegraphics[width=\textwidth]{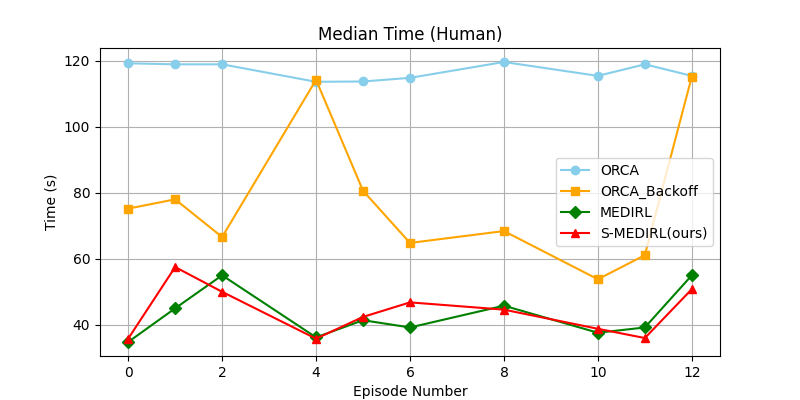} 
        \caption{Time taken by the human to complete the episodes with deadlock}
        \label{fig:human_time}
    \end{subfigure}
    
    \begin{subfigure}{\columnwidth} 
    \centering
        \includegraphics[width=\textwidth]{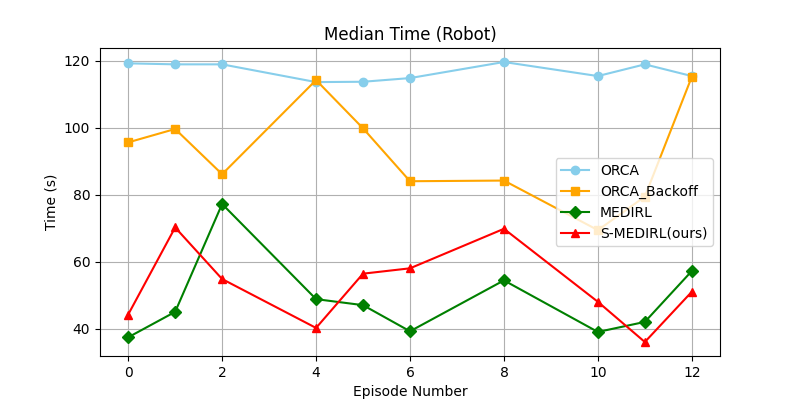} 
        \caption{Time taken by the robot to complete the episodes with deadlock.}
        \label{fig:robot_time}
    \end{subfigure}

    \caption{The figure depicts the overall time taken by the robot and the time taken to reach their respective goals. For clarity, we only include the cases where deadlock happens, so the yielding action by the robot is required.  Fig.~\ref{fig:human_time} depicts the time taken by the human to complete the task when the robot is operating using different algorithms. We note that while ORCA baseline fails for all these and times out at about 120 seconds. The ORCA backoff agent succeeds for all but two cases, reported in Fig.~\ref{fig: success}, but in general takes about 20 seconds extra for the human and 30 seconds extra for the robot.  (Fig.~\ref{fig:robot_time})}
    \label{fig: time}
\end{figure}
Quantitatively, we evaluate the quality of the trajectory of our agent by looking at the median time taken by the human and robot to complete the task over the 5 runs on each episode.  Fig.~\ref{fig: time} describes the time taken for both the human and the robot to complete their task, i.e., reach their respective goals for navigating all episodes with deadlock.
As is seen for the success metrics in Fig.~\ref{fig: success}, the ORCA agent fails to complete the task; this is reported in Fig.~\ref{fig: time} as the ORCA agent times out at 2 minutes of task completion time. The ORCA backoff agent can complete the task; however, it takes about 30 seconds longer on average than the IRL agents, as the yielding position is arbitrary.
\section{Conclusion}
\label{sec:conclusion}
This paper presents a MEDIRL-based agent for navigating indoor environments with complex scene geometry with humans. The proposed agent is trained on top-down images of the scene, the agent's trajectories, human velocity, heading, and robot goal. Training on this image data takes away the reliance on handcrafted features that can be arbitrary to different robots' tasks and design preferences. The agent learns from human demonstration how to yield to oncoming traffic and go forward through a narrow crossing. We compare the performance of the agent with that of a nominal ORCA agent and an ORCA-backoff agent. We saw that the IRL agents were implicitly able to learn to avoid deadlock and could complete the task faster than the baseline comparisons. To improve the agent performance when it gets stuck in unseen parts of the scene, we propose a way to extrapolate demonstration data and a smoothing loss to eliminate aberrations in training the reward. The S-MEDIRL shows comparable task completion times with improved success rate of all tasks as compared to the MEDIRL agent. 
We emphasize that while it is hard to account for desired social behavior, we may want the agent to exhibit that learning from demonstration can help us learn these norms and goals end-to-end. 
\section{Discussion}
\label{sec:discussion}

The purpose of this study is to shed light on the need to learn social behavior from demonstrations. We describe how social behavior is implicitly a part of our trajectory planning as humans and is hard to model in an exhaustive list for social agents to be able to accomplish in a rule-based manner. We show that learning this behavior from the demonstration of raw image data removes any design biases that may come from handcrafted feature design for the completion of specific tasks. We never explicitly work on detecting or avoiding deadlock and observe that the agent can implicitly learn deadlock avoidance behavior from demonstration data. However, like most agents that rely on demonstrations for learning motion, we rely heavily on there being good demonstrations available. We also find that this qualitative good metric is hard to quantify and differs for different tasks. Previous approaches have fixed what these qualitative goodness metrics are and account for preference learning. Our approach learns these biases, assuming good demonstration data is available. 

Our future work wants to extend our approach to encode nominal goodness metrics into our system design. We want to use a Vision and Language model to predict the goodness of a demonstration and then encode these preferences into our agent as a trajectory rank similar to \cite{gan2022energy}.

\section*{Acknowledgment}
{
We would like to thank our collaborators at the University of Michigan Museum of Art, Grace VanderVliet and John Turner, for their constant support through our progress. }

{\footnotesize
\balance
\bibliographystyle{bib/IEEEtran}
\bibliography{bib/IEEEabrv,bib/strings-abrv,bib/ieee-abrv,bib/refs}
}

\end{document}